\documentclass[11pt]{article}

\usepackage[english]{babel}

\usepackage[marginratio=1:1,height=600pt,width=460pt,tmargin=100pt]{geometry}

\usepackage{amsmath}
\usepackage{graphicx}
\usepackage[utf8]{inputenc} 
\usepackage[T1]{fontenc}    
\usepackage[hidelinks, colorlinks=true,citecolor=blue]{hyperref} 
\usepackage{url}
\usepackage{booktabs}
\usepackage{amsfonts}
\usepackage{nicefrac}
\usepackage{microtype}
\usepackage{xcolor}
\usepackage[normalem]{ulem}
\usepackage{amsmath}
\usepackage{amssymb}
\usepackage{color}
\usepackage[numbers]{natbib}
\usepackage{caption}
\usepackage{subcaption}
\usepackage{bm}
\usepackage{bbm}
\usepackage[flushleft]{threeparttable}
\usepackage[shortlabels]{enumitem}
\usepackage{wrapfig}
\setlist{leftmargin=6mm}
\usepackage{multirow}
\usepackage{titlesec}
\usepackage{setspace}

\usepackage{mathtools, mathrsfs}
\usepackage{textcomp, gensymb}
\usepackage{dsfont}
\usepackage{soul}

\usepackage{algorithm}
\usepackage{algorithmic}
\newlength\myindent
\usepackage{amsthm}

\newtheorem{conjecture*}{Conjecture}
\theoremstyle{remark}

\allowdisplaybreaks

\title{GreenTEA: Gradient Descent with Topic-modeling and Evolutionary Auto-prompting}

\author{
  Zheng Dong, Luming Shang, and Gabriela Olinto
}

\date{\small
    Amazon Buyer Risk Prevention
}

\begin{document}
\maketitle

\begin{abstract}
High-quality prompts are crucial for Large Language Models (LLMs) to achieve exceptional performance. However, manually crafting effective prompts is labor-intensive and demands significant domain expertise, limiting its scalability. Existing automatic prompt optimization methods either extensively explore new prompt candidates, incurring high computational costs due to inefficient searches within a large solution space, or overly exploit feedback on existing prompts, risking suboptimal optimization because of the complex prompt landscape. To address these challenges, we introduce \texttt{GreenTEA}, an agentic LLM workflow for automatic prompt optimization that balances candidate exploration and knowledge exploitation. It leverages a collaborative team of agents to iteratively refine prompts based on feedback from error samples. An analyzing agent identifies common error patterns resulting from the current prompt via topic modeling, and a generation agent revises the prompt to directly address these key deficiencies. This refinement process is guided by a genetic algorithm framework, which simulates natural selection by evolving candidate prompts through operations such as crossover and mutation to progressively optimize model performance. Extensive numerical experiments conducted on public benchmark datasets suggest the superior performance of \texttt{GreenTEA} against human-engineered prompts and existing state-of-the-arts for automatic prompt optimization, covering logical and quantitative reasoning, commonsense, and ethical decision-making.

\end{abstract}

\section{Introduction}
The performance of LLMs is highly dependent on the quality of the prompts. Manual prompt engineering \citep{brown2020language, kojima2022large}, though widely adopted, requires substantial human effort and domain expertise, posing challenges when adapting to diverse tasks. To mitigate this dependency, automatic prompt optimization has emerged as a promising direction \citep{ramnath2025systematic}. Early methods primarily adopt discrete search strategies \citep{jiang2020can, zhoularge} by enumerating a wide range of prompt candidates, enabling extensive exploration but often incurring high computational costs due to inefficient search dynamics. An alternative line of research focuses on feedback-driven refinement \citep{dong2024pace, prasad2023grips, sun2023autohint, zhang2023tempera} by learning from error samples and improve existing prompts in an iterative way. While effective in exploiting learned knowledge, these approaches are susceptible to suboptimal convergence in the presence of complex and non-convex landscapes in prompt solution space. These limitations highlights the necessity of a well-designed optimization algorithm for navigating the complex prompt space in a cost-efficient way.

In this paper, we introduce \texttt{GreenTEA}, an agentic workflow for automatic prompt optimization that emphasizes both effectiveness and efficiency. \texttt{GreenTEA} adopts a feedback-driven, iterative framework in which one LLM agent analyzes the deficiencies of current prompts and another generates improved candidates based on the identified deficiencies. To guide this process, we introduce a topic-modeling mechanism that clusters error samples by the current prompts. This facilitates the identification of major error patterns and ensures that the collected samples sent to the analyzing agent are semantically coherent. New prompts are generated via a novel gradient-guided genetic algorithm (GA), which produces a population of candidates through evolutionary operations on existing prompts, guided by the feedback. This design draws on the success of evolutionary algorithms in prompt optimization \citep{baumann2024evolutionary, guoconnecting, xu2022gps}, striking a balance between exploration and exploitation within the prompt space. The main contributions of this paper include:
\begin{itemize}
    \item We present \texttt{GreenTEA}, an agentic workflow for automatic prompt optimization that emphasizes both effectiveness and efficiency.
    
    \item We introduce an error topic modeling mechanism to improve the optimization efficiency by extracting the major error and collecting feedback based on semantically coherent samples.

    \item We develop a gradient-guided evolutionary algorithm that enhances the robustness of prompt optimization by balancing exploration and exploitation in the prompt search space.

    \item Comprehensive experiments show that \texttt{GreenTEA} consistently achieves expert-level prompts across different scenarios.

\end{itemize}

\section{Related works}

\noindent \textbf{LLMs for automatic prompt optimization.}
Early work regards the prompt as a learnable artifact and leverages LLMs to serve as black–box optimizer \citep{zheng2023can} to solve the prompt optimization problem.
Various approaches adopt self‑refinement via LLMs to critique answers and regenerate the prompts iteratively \citep{pryzant2023automatic, yang2024large}.
Recent work of DSPy \citep{khattab2023dspy} generalizes this idea into a compiler that represents an unified reasoning and optimization pipeline.
These studies establish the premise that prompt optimization can be automated by exploiting the existing knowledge, yet they adopt generic random or greedy edits that may potentially
lead to local optima \citep{guoconnecting}.

\vspace{0.1in}
\noindent \textbf{Evolutionary prompt search.}
Evolutionary algorithms (EAs) have emerged as a powerful tool for automatic prompt optimization by iteratively refining a \emph{population} of candidate solutions through biologically inspired operators. Prior studies \citep{baumann2024evolutionary, liu2023algorithm} demonstrate the effectiveness of LLM-based operators, while EVOPROMPT \citep{guoconnecting} presents a generic framework that couples LLMs with EAs, enabling a broad family of EA variants within a unified pipeline. Recent work of GAAPO \citep{secheresse2025gaapo} compares different evolution strategies in a modularized framework via comprehensive evaluations. In contrast to these approaches, our \texttt{GreenTEA} integrates a gradient‑guided exploration mechanism that directs evolution explicitly towards the major failure modes.

\vspace{0.1in}
\noindent \textbf{Gradient‑guided genetic algorithm.}
Genetic algorithm (GA) \citep{holland1992adaptation} is a well‑established class of evolutionary search methods and have been widely adopted for prompt optimization \citep{tanaka2023genetic, xu2022gps}. They balance exploration with exploitation via fitness‑based selection, achieved through crossovers and mutations that traverse the vast prompt space. Direction‑guided variants further accelerate convergence by biasing mutations with gradient rather than random perturbations, consistently achieving enhanced performance across domains such as protein–ligand docking \citep{guan2016edga}, system‑fairness testing \citep{fan2022explanation}, and molecular generation \citep{zhuang2025gradientga}. While they remain largely unexplored for prompt evolution, our \texttt{GreenTEA} closes that gap by embedding a topic‑guided gradient into the GA loop, yielding faster convergence and more interpretable evolutionary trajectories.

\section{Methodology: GreenTEA}

Given a LLM $\mathcal{M}$, the goal of prompt optimization is to find the best-performing prompt $\mathcal{P}^*$ on a training set $\mathcal{D}_{\rm{tr}} = \{(q_1, a_1), \dots, (q_N, a_N)\}$ under a performance metric $s$, \textit{i.e.}, $\mathcal{P}^* = {\rm argmax}_{\mathcal{P} \in \Omega}\ s(\mathcal{P}, \mathcal{D}_{{\rm tr}}, \mathcal{M})$. Here $(q_i,a_i)$ represents the $i$-th question-answer pair. This optimization problem is highly non-convex and intractable. In the following, we introduce our \texttt{GreenTEA} framework that use LLM agents to find the approximate solution in an iterative way.

\subsection{Overview: Iterative prompt optimization}

\begin{figure}[!t]
\centering
\begin{subfigure}[h]{1.0\linewidth}
\includegraphics[width=\linewidth]{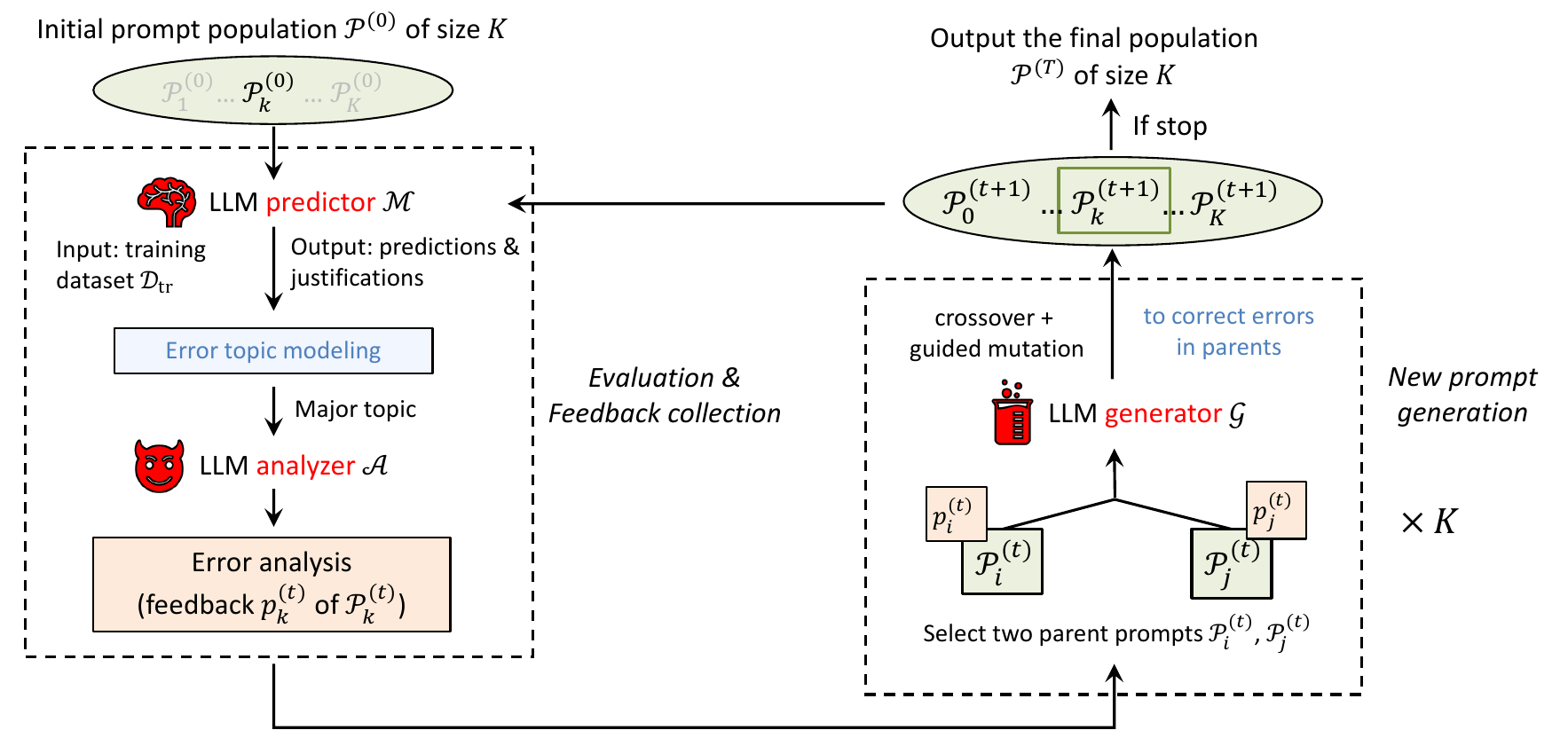}
\end{subfigure}
\caption{GreenTEA framework. The first stage (left) evaluates the prompts on the predictor $\mathcal{M}$ and collect feedback on them via the analyzer $\mathcal{A}$. The second stage (right) generates new prompts via the generator $\mathcal{G}$. The two stages are performed iteratively until termination criteria are met.}
\label{fig:greentea-framework}
\end{figure}

The framework contains a feedback-collection stage and a prompt generation stage that performed in an iterative loop to progressively enhance the prompt, as illustrated in Figure~\ref{fig:greentea-framework}. The algorithm starts with a initial population of prompts $\mathcal{P}^{(0)}$ of size $K$. These prompts are simple and generic without being fine-crafted for the task. 

In the first stage of $t$-th iteration, each prompt $\mathcal{P}_k^{(t)}$ in the existing population $\mathcal{P}^{(t)}$ will be evaluated on $\mathcal{D}_{\rm{tr}}$ using the LLM predictor $\mathcal{M}$. We then apply the topic modeling to all the cases that $\mathcal{M}$ have made the wrong predictions on to group them into different topic clusters. Therefore, cases in the same cluster are semantically closer with each other than those in different clusters. We collect those in the cluster with the largest size and input them to the second agent, named the LLM analyzer $\mathcal{A}$. The analyzer summarizes the mistakes and outputs an error analysis $p_k^{(t)}$ as the feedback of $\mathcal{P}_k^{(t)}$ that pinpoints the key deficiencies of the prompt.

After the feedback collection, the second stage contains the third agent, named the LLM generator $\mathcal{G}$, that produce new prompts to address the deficiencies of the existing prompt. We use a gradient-guided GA to evolute the prompts. Instead of optimizing a single prompt, the generator $\mathcal{G}$ takes two existing prompts (the parents) as the input, combine them into one, and revise the prompt based on the feedback of the parents to generate a new prompt (the child). The generator $\mathcal{G}$ will perform the generation by $K$ times to have the new population $\mathcal{P}_k^{(t+1)}$ of size $K$, which will be evaluated and further refined in the next iteration. This cycle continues until predefined termination criteria are met. This evolutionary procedure enhances the exploratory ability of the framework and avoids the optimization trajectory being trapped in the local minima.

The output of \texttt{GreenTEA} is the prompt population $\mathcal{P}^{(T)}$ being optimized over $T$ iterations. See Appendix~\ref{app:GT-algorithm} for algorithm details.
In the following, we elaborate on the topic modeling for collecting wrong predictions and the guided GA for prompt generation.

\subsection{Topic modeling for feedback collection}
We observe diverse reasons behind the wrong predictions by the predictor $\mathcal{M}$. To make it easier for the analyzer $\mathcal{A}$ to summarize the error and identify the prompt deficiency, we adopt the topic modeling to guide the selection of wrong prediction samples. Assuming we have the output of $\mathcal{M}$ on each question $q_i$, denoted as $\widehat{a_i} \sim g_{\mathcal{M}}(\cdot | \mathcal{P}_k^{(t)}, q_i)$. Here $g_{\mathcal{M}}$ represents the text generation mechanism of $\mathcal{M}$ that returns the most likely textual output given the input of the prompt and the question.
We collect wrong prediction samples, denoted by $\mathcal{W} = \{(q_i, a_i, \widehat{a_i}): a_i \neq \widehat{a_i}\}$, and perform clustering on $\mathcal{W}$ based on the textual embedding of each $a_i$. Both the choices of the text embedding model and the clustering algorithm are flexible. For instance, we apply a pretrained sentence BERT \citep{reimers2019sentence} to create the embedding and the K-nearest neighbors for clustering. 
Samples in the largest cluster are retrieved as the final input to the analyzer $\mathcal{A}$ for error analysis, each including $q_i, a_i$, and $\widehat{a_i}$.
By doing so, the picked samples are semantically close, allowing for the analyzer to focus on the major type of error during its analysis. 


The output of the analyzer $\mathcal{A}$ is an analysis of the error that $\mathcal{M}$ has made by comparing the model prediction $\widehat{a_i}$ with the true answer $a_i$. The error analysis contains a guidance on how the existing prompt can be improved to address its deficiencies and output the correct predictions. Note that we make the improvement guidance not question-specific but as general as possible to be able to applied to similar questions that have not been encountered. 

\subsection{Prompt generation via guided GA}

Given the current population $\mathcal{P}^{(t)}$, the guided GA generates a new prompt via the following steps:
\begin{itemize}
    \item \textbf{Parent selection}. We use roulette wheel selection method \citep{lipowski2012roulette} to select two parent prompts $\mathcal{P}_i^{(t)}$ and $\mathcal{P}_j^{(t)}$ based on their fitness scores.
    The fitness score $f_k$ for each prompt $\mathcal{P}_k^{(t)}$ is measured by the scoring function on the training data, \textit{e.g.,} $f_k = s(\mathcal{P}_k^{(t)}, \mathcal{D}_{\rm tr}, \mathcal{M})$. 
    The probability of selecting the $k$-th prompt as a parent is $e_k = f_k/\sum_{k'=1}^{K} f_{k'}$.
    
    \item \textbf{Child generation}. The LLM generator $\mathcal{G}$ takes the selected parent prompts together with their feedback as the input, \textit{e.g.}, $\{\mathcal{P}_i^{(t)}, \mathcal{P}_j^{(t)},  p_i^{(t)}, p_j^{(t)}\}$, and generates a child prompt $\mathcal{P}_k^{(t+1)}$. The generator performs this prompt evolution by i) a crossover to combine the parent prompts into an child prompt that inherits their traits; ii) a mutation to introduce modification to the child prompt. We use corresponding instructions to guide the mutation to incorporate feedback into the child prompt.
\end{itemize}

We repeat the above steps by $K$ times to produce $K$ child prompts. These child prompts will be evaluated on $\mathcal{D}_{\rm tr}$ to get their fitness score. We then merge them with the current population and retain the top $K$ prompts with the highest fitness scores to form the updated population $\mathcal{P}^{(t+1)}$. This process ensures the improvement of the population’s overall quality, concluding with the best prompt in the updated population being designated as the optimal prompt.

\section{Experiments}

We present the numerical results of prompt optimization that demonstrate the superior performance of \texttt{GreenTEA} across different tasks. 



\subsection{Experiment setup}

We conduct experiments on the following datasets, including:
\begin{itemize}
    \item \textbf{GSM8K} \citep{cobbe2021training}: A dataset of 8.5K grade-school math word problems designed to evaluate quantitative reasoning and step-by-step problem solving. Each instance includes a problem description and a detailed, free-form answer explanation.

    \item \textbf{ETHOS} \citep{mollas2022ethos}: An English hate speech detection dataset containing 997 online comments labeled for attributes such as hate speech, discrimination, and offensive language. It evaluates a model's capability in ethical and socially sensitive reasoning.

    \item \textbf{PIQA} \citep{bisk2020piqa}: A physical commonsense reasoning benchmark with 16,000 multiple-choice questions. Each question describes a real-world scenario, and the model must select the more plausible of two possible solutions.
    
    \item \textbf{BBH (Big-Bench Hard)} \citep{suzgun2023challenging}: A challenging subset of the BIG-Bench benchmark consisting of 23 diverse tasks requiring multi-step logical reasoning, such as date manipulation, causal judgment, and word sorting. Each task includes hundreds of examples with high variance in difficulty. 
\end{itemize}

The choice of the agents in \texttt{GreenTEA} is flexible and can be either open‑ or closed‑source. While the predictor $\mathcal{M}$ is picked to satisfy the deployment constraints of a given application (latency, cost, privacy), we recommend using a more capable LLM for both the analyzer $\mathcal{A}$ and the generator $\mathcal{G}$ \citep{khattab2023dspy, yang2024large}, as larger models possess richer world knowledge and stronger reasoning abilities to yield gains when optimizing the prompts. In the following experiments, we use closed-source models from the Anthropic Claude family, choosing the predictor to be \textit{Claude 3 Sonnet}, and the analyzer and the generator to be \textit{Claude 3.5 Sonnet}. See more details of experimental setup and prompt templates in Appendix~\ref{app:experimental-setup}.

\subsection{Comparison with baselines}

\begin{wraptable}{r}{.65\linewidth}
\vspace{-0.15in}
  \caption{Testing accuracy of \texttt{GreenTEA} and baselines}
  \vspace{-0.1in}
  \centering
  \resizebox{1.\linewidth}{!}{
    \centering
    \begin{tabular}{l||ccccc}
    \toprule
    \toprule
    \textbf{Method} & \multicolumn{1}{c}{\textbf{GSM8K}} & \multicolumn{1}{c}{\textbf{ETHOS}} & \multicolumn{1}{c}{\textbf{PIQA}} & \multicolumn{1}{c}{\textbf{BBH}} \\
    \midrule
    CoT prompt & $82.43_{(1.19)}$ & $78.33_{(1.25)}$ & $83.97_{(0.72)}$ & $77.48_{(2.05)}$ \\
    \texttt{OPRO} \citep{yang2024large} & $88.10_{(0.93)}$ & $83.33_{(2.05)}$ & $86.05_{(0.61)}$ & $79.58_{(1.26)}$\\
    \texttt{ProTeGi} \citep{pryzant2023automatic} & $84.63_{(1.02)}$ & $\textbf{85.67}_{(1.49)}$ & $85.72_{(0.84)}$ & $78.86_{(1.77)}$ \\
    \texttt{EVOPROMPT} \citep{guoconnecting} & $86.27_{(0.58)}$ & $82.00_{(1.41)}$ & $86.43_{(0.36)}$ & $80.12_{(1.43)}$ \\
    \midrule
    \texttt{GreenTEA} & $\textbf{91.37}_{(0.45)}$ & $84.67_{(1.89)}$ & $\textbf{89.86}_{(0.28)}$ & $\textbf{82.23}_{(1.13)}$ \\
    \bottomrule
    \bottomrule
  \end{tabular}
  }
  \begin{tablenotes}
  \item {\footnotesize *Numbers in parentheses are standard errors for three independent runs.}
  \end{tablenotes}
  \label{tab:performance-comparison}
  \vspace{-0.1in}
\end{wraptable}

We compare our \texttt{GreenTEA} against three state-of-the-arts in automatic prompt optimization that have methodological relations to \texttt{GreenTEA}: (i) \texttt{OPRO} \citep{yang2024large} leverages LLMs as optimizer to find instructions that maximize the model performance; (ii) \texttt{ProTeGi} \citep{pryzant2023automatic} exploits feedback on the existing prompt and use textual gradient descent to optimize the prompt; (iii) \texttt{EVOPROMPT} \citep{guoconnecting} adopts an evolutionary framework for prompt optimization without using the prompt feedback. We also evaluate the performance of chain-of-thoughts (CoT) prompts \citep{kojima2022large} and compare it with \texttt{GreenTEA}.

Table~\ref{tab:performance-comparison} reports the accuracy of the predictor $\mathcal{M}$ equipped with different prompts on the testing set on different tasks. The prompts optimized by \texttt{GreenTEA} achieve a consistently better or comparable performance against those from other prompt optimization baselines. The \texttt{GreenTEA} also significantly improve the prompt performance from the manually designed CoT prompts across the evaluated scenarios.

\subsection{Ablation study}

We perform an ablation study to demonstrate the effectiveness of the topic-modeling and gradient-guided GA in prompt optimization. Specifically, we compared two variants of \texttt{GreenTEA} that have these key components removed:
\begin{itemize}
    \item \texttt{GreenTEA-TM-GGA}: this variant uses a classic GA instead of the gradient-guided GA to generate child prompts. The LLM generator $\mathcal{G}$ performs random mutation by simply rephrasing the intermediate prompt from crossover of two parent prompts, without deliberately addressing specific errors. Note that we do not need to perform feedback collection, thus topic modeling is also excluded in this variant.

    \item \texttt{GreenTEA-TM}: this variant uses the gradient-guided GA but replaces the topic modeling with random sampling, collecting incorrect samples by randomly selecting a subset of wrong predictions from the available pool.
\end{itemize}

\begin{figure}[!t]
\centering
\begin{subfigure}[h]{0.48\linewidth}
\includegraphics[width=.95\linewidth]{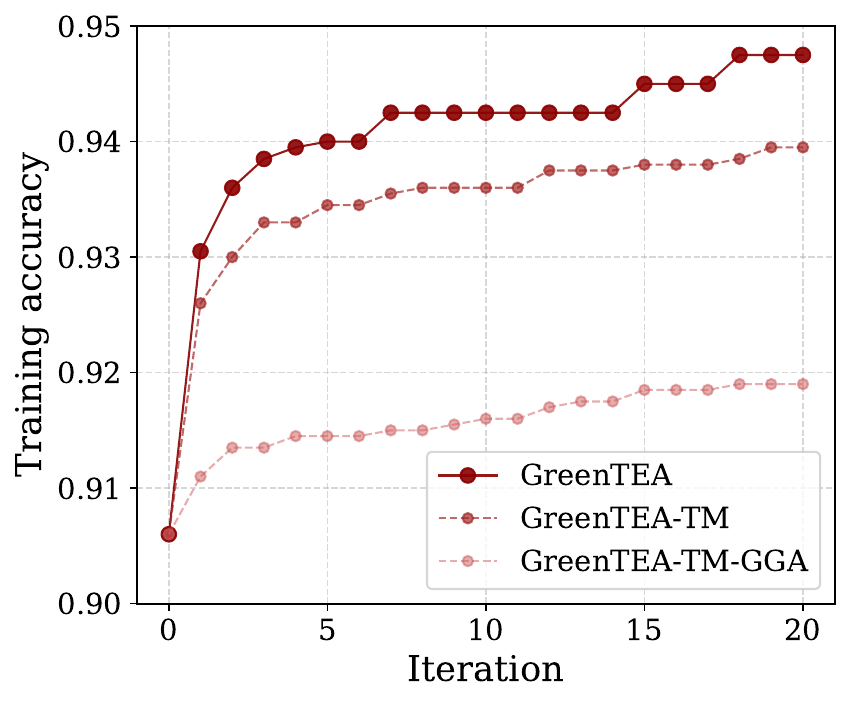}
\caption{GSM8K}
\end{subfigure}
\begin{subfigure}[h]{0.48\linewidth}
\includegraphics[width=.95\linewidth]{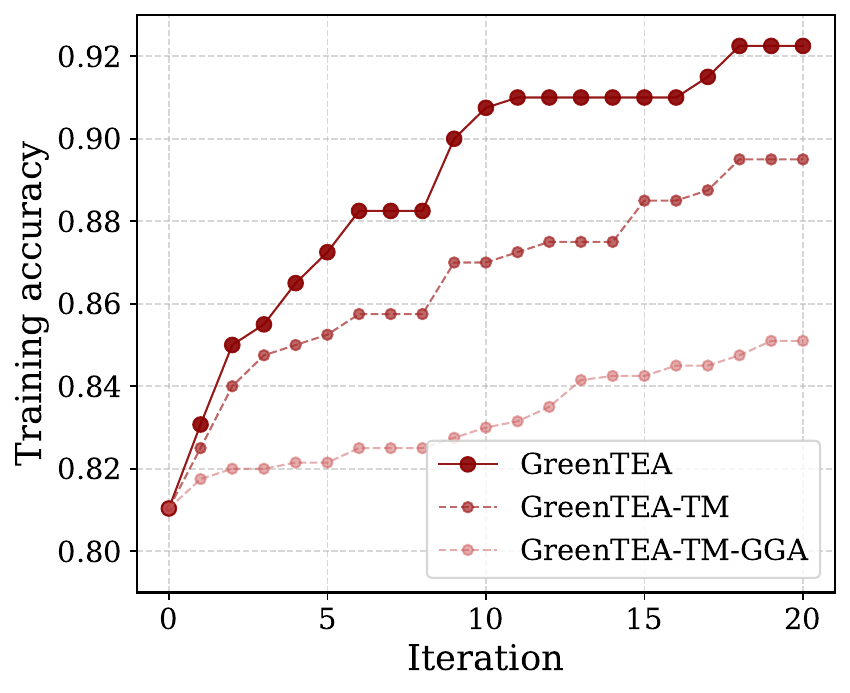}
\caption{ETHOS}
\end{subfigure}
\vspace{-0.05in}
\caption{Training accuracy evolution by \texttt{GreenTEA} and two ablated variants on GSM8K and ETHOS datasets.}
\label{fig:ablation-study}
\end{figure}

To better investigate the performance difference between variants, we visualize the evolution of the average model accuracy over the whole population on the training set of GSM8K and ETHOS datasets after each iteration in Figure~\ref{fig:ablation-study}. 
We observe that the complete \texttt{GreenTEA} consistently achieves higher accuracy and faster convergence compared to both ablated variants, as it rapidly improves within the first few iterations and continues to exhibit steady gains. The performance drop in \texttt{GreenTEA-TM} highlights the benefits of topic modeling in improving the optimize efficiency as it facilitates more targeted feedback learning. The earlier plateau and lower final accuracies of \texttt{GreenTEA-TM-GGA} further underscore the significance of gradient-guided evolution in prompt optimization.

\begin{figure}[!t]
\centering
\begin{subfigure}[h]{0.98\linewidth}
\includegraphics[width=1.\linewidth]{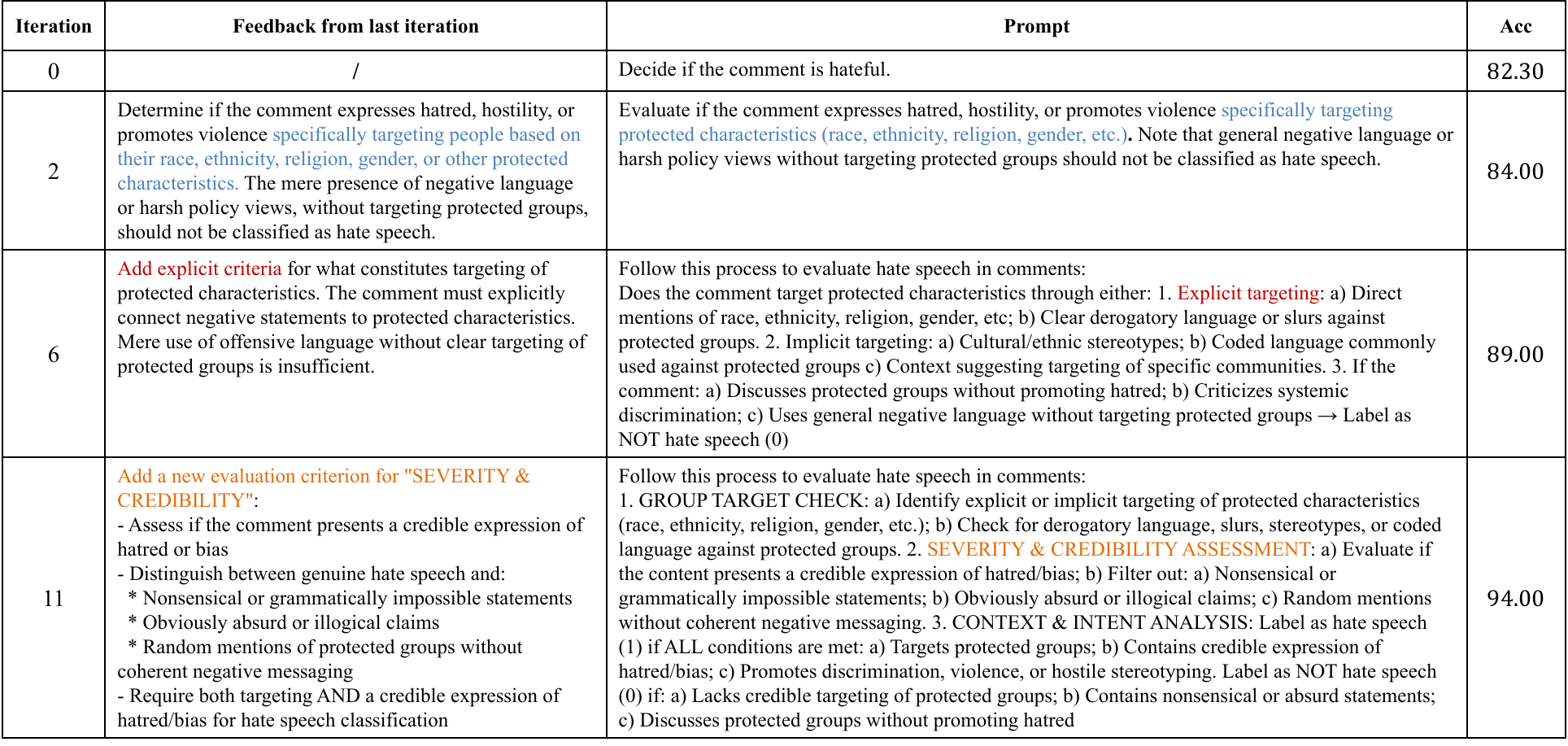}
\end{subfigure}
\caption{Prompt evolution on ETHOS dataset. Key messages in the feedback and the corresponding adjustments in the prompts at each iteration are highlighted in the same color.}
\label{fig:prompt-evolution}
\end{figure}


\subsection{Model interpretation}

To illustrate how \texttt{GreenTEA} systematically enhances prompt quality, Figure~\ref{fig:prompt-evolution} presents the feedback and corresponding optimized prompts collected across three iterations for the ETHOS dataset. At each iteration, the LLM analyzer $\mathcal{A}$ identifies key deficiencies in the prompt—such as lack of clarity on group targeting, missing criteria for hate speech, or ambiguity in assessing intent and severity. These critical insights from the LLM analyzer $\mathcal{A}$ (highlighted in color) are explicitly integrated into the prompt by the LLM generator $\mathcal{G}$. Beginning with a basic prompt like ``Decide if the comment is hateful'', the \texttt{GreenTEA} with the backbone of Claude model family progressively refines instructions into detailed, multi-step guidelines. This structured refinement effectively guides the LLM predictor $\mathcal{M}$ toward more accurate classifications. These improvements of the prompts directly correlate with substantial accuracy gains at iterations $2$, $6$, and $11$ in Figure~\ref{fig:ablation-study}(b), demonstrating \texttt{GreenTEA}'s capability in optimizing more interpretable and high-performing prompts.

\section{Discussion}

We propose \texttt{GreenTEA}, an agentic workflow for prompt optimization that emphasizes both effectiveness and efficiency. It iteratively refines prompts by incorporating feedback from previous performance and imitating the natural evolutionary process to generate stronger solutions. Numerical experiments show the superior performance of \texttt{GreenTEA} against current state-of-the-arts.
An ablation study further demonstrates the benefits of topic modeling and guided genetic algorithms in enhancing the efficiency and robustness of the optimization process in complex prompt solution space.
Extensions of \texttt{GreenTEA} include adopting other clustering methods, \textit{e.g.}, spectral clustering that has demonstrated strong empirical performance \citep{liu2025network} to identify informative error patterns; incorporating advanced scoring metrics beyond aggregated accuracy to capture key features that reflect prompt quality \citep{yang2024large}; designing efficient workflows \citep{kim2024llm} to manage the growing complexity of prompts over iterations; and leveraging its non-parametric optimization capabilities for broader solution optimization tasks, \textit{e.g.}, high-dimensional event modeling \citep{cheng2025deep, dong2023deep, dong2023non, dong2024spatio, dong2023spatiotemporal, dong2025conditional, zhu2022neural}, that go beyond prompt refinement.

\bibliographystyle{apalike}
\bibliography{arxiv_refs}

\newpage
\appendix

\section{Algorithm}
\label{app:GT-algorithm}

we present the full algorithmic workflow of \texttt{GreenTEA} in Algorithm~\ref{alg:CEG-data-generation}. The method maintains a population of prompts that are iteratively refined through a feedback-guided evolutionary process. In each iteration, error patterns are extracted from model predictions via topic modeling, analyzed by an LLM agent, and used to guide new prompt generation through crossover and mutation.

\begin{algorithm}[!htb]
\begin{algorithmic}
    \STATE {\bfseries Input:} Population size $K$, max number of iteration $T$, scoring function $s(\cdot)$, training data $\mathcal{D}_{\rm tr}$, LLM predictor $\mathcal{M}$, LLM analyzer $\mathcal{A}$, LLM generator $\mathcal{G}$\;
    \STATE {\bfseries Initialization:} Population $\mathcal{P}^{(0)} = \{\mathcal{P}_0^{(0)}, \dots, \mathcal{P}_{k-1}^{(0)}\}$, $t=0$\; 
    \WHILE{$t\leq T$}
        \FOR{each prompt $\mathcal{P}_k^{(t)}$ in $\mathcal{P}^{(t)}$ that has not been evaluated}
         \STATE 1. Get score $f_k = s(\mathcal{P}_k^{(t)}, \mathcal{D}_{\rm tr}, \mathcal{M})$ and collect $\{\widehat{a_n}\}_{n=1}^{N}$;
         \STATE 2. Get wrong samples $\mathcal{W}_{k}^{(t)}$ using topic modeling;
         \STATE 3. Input $\mathcal{W}_{k}^{(t)}$ to $\mathcal{A}$ and obtain feedback $p_k^{(t)}$ of $\mathcal{P}_k^{(t)}$;
        \ENDFOR
        \STATE $\mathcal{P}^{(t)} = $ prompts in $\mathcal{P}^{(t)}$ with the top-$K$ scores;
        \STATE $\mathcal{P}^{(t+1)} = \mathcal{P}^{(t)}$;
        \FOR{$k \in \{0, \dots, K-1\}$}
         \STATE 1. Select parent prompts $\mathcal{P}_i^{(t)}, \mathcal{P}_j^{(t)}$ from $\mathcal{P}^{(t)}$ using roulette wheel selection;
         \STATE 2. Generate a new prompt $\mathcal{P}_k^{(t+1)}$ by inputting $\mathcal{P}_i^{(t)}, \mathcal{P}_j^{(t)}, p_i^{(t)}, p_j^{(t)}$ to the generator $\mathcal{G}$;
         \STATE 3. $\mathcal{P}^{(t+1)} = \{\mathcal{P}^{(t+1)}, \mathcal{P}_k^{(t+1)}\}$
        \ENDFOR
        \STATE $t \leftarrow t+1$;
    \ENDWHILE
    \RETURN $\mathcal{P}^{(T)} = $ prompts in $\mathcal{P}^{(T)}$ with the top-$K$ scores;
\end{algorithmic}
\caption{\texttt{GreenTEA} for automatic prompt optimization}
\label{alg:CEG-data-generation}
\end{algorithm}

\section{Details of experimental setup}
\label{app:experimental-setup}

\textbf{Dataset.} The testing set contain 1,000 samples for GSM8K and PIQA data, contain 100 samples for ETHOS data, and contain 25 samples for each BBH task. 

\begin{figure}[!t]
\centering
\begin{subfigure}[h]{0.9\linewidth}
\includegraphics[width=1.\linewidth]{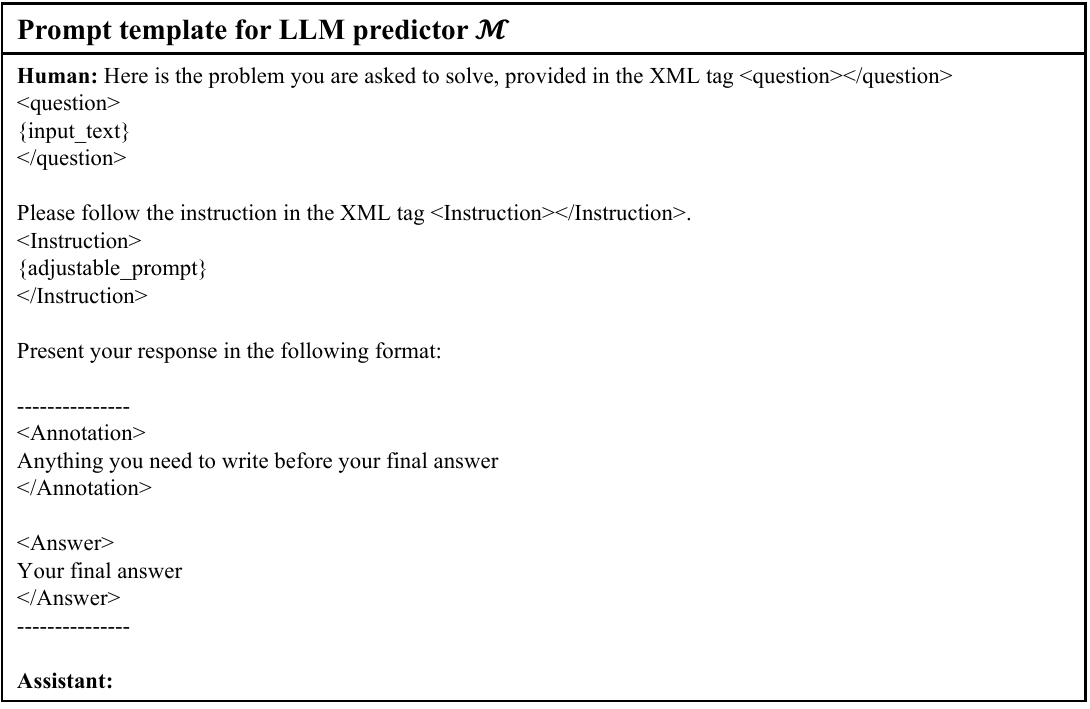}
\end{subfigure}
\caption{Prompt template for predictor $\mathcal{M}$}
\label{fig:predictor-template}
\end{figure}

\vspace{0.1in}
\noindent \textbf{Experiment configuration.} We set the temperature of the LLM predictor $\mathcal{M}$ to be $0.0$ when evaluating the performance of generated prompts, and set the default temperature to be $1.0$ for LLM analyzer $\mathcal{A}$ and generator $\mathcal{G}$ to generate diverse outputs. For the generic algorithm, we set the number of iteration to be $20$ and the population size to be $4$, \textit{i.e.}, generate $4$ new prompts in each iteration and keep the best $4$ prompts as the updated population after each iteration. We evaluate the testing accuracy of the prompts in the last population and report their average as the final prompt accuracy by \texttt{GreenTEA}. The results of the testing accuracy in Table~\ref{tab:performance-comparison} are averaged over three independent runs.

\begin{figure}[!t]
\centering
\begin{subfigure}[h]{0.9\linewidth}
\includegraphics[width=1.\linewidth]{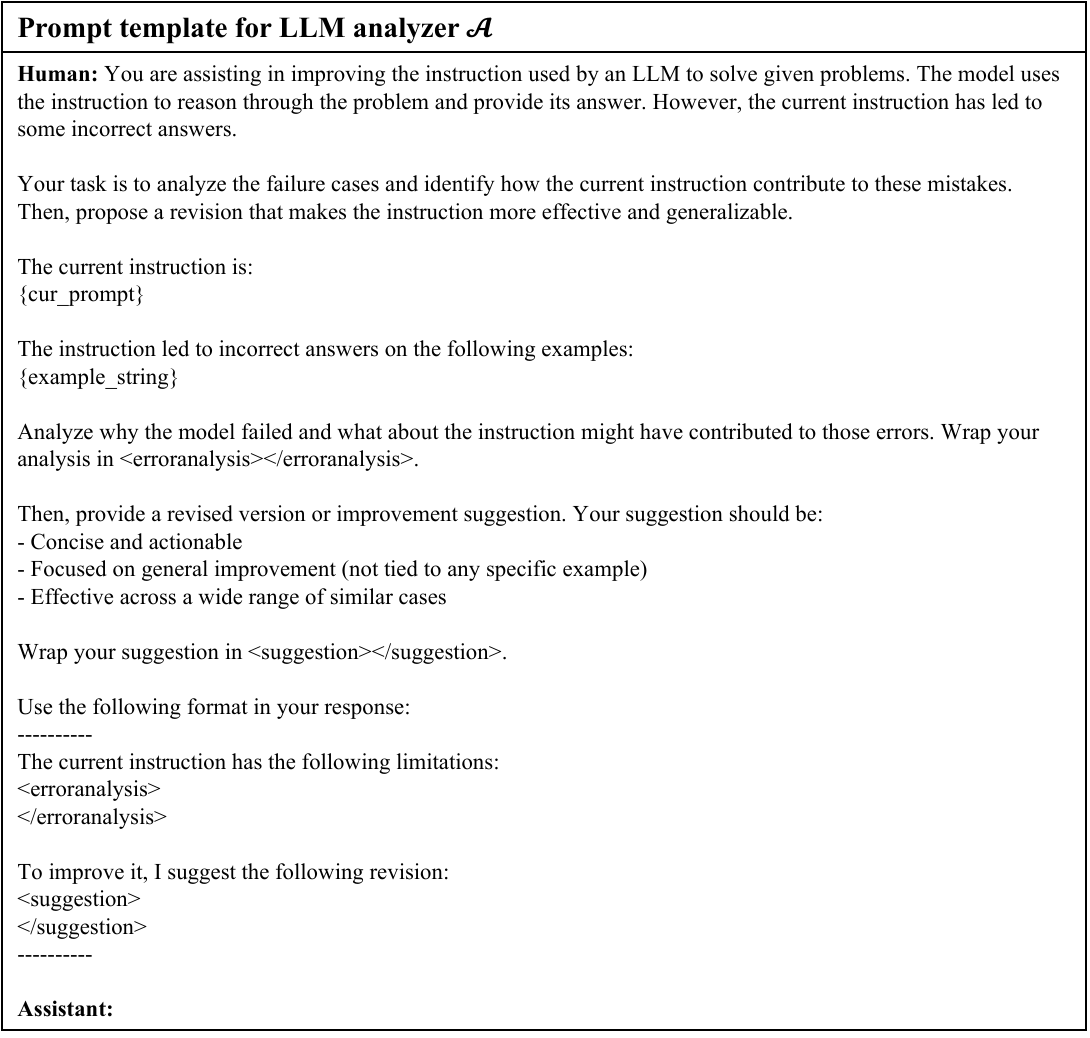}
\end{subfigure}
\caption{Prompt template for analyzer $\mathcal{A}$}
\label{fig:analyzer-template}
\end{figure}

\begin{figure}[!t]
\centering
\begin{subfigure}[h]{0.9\linewidth}
\includegraphics[width=1.\linewidth]{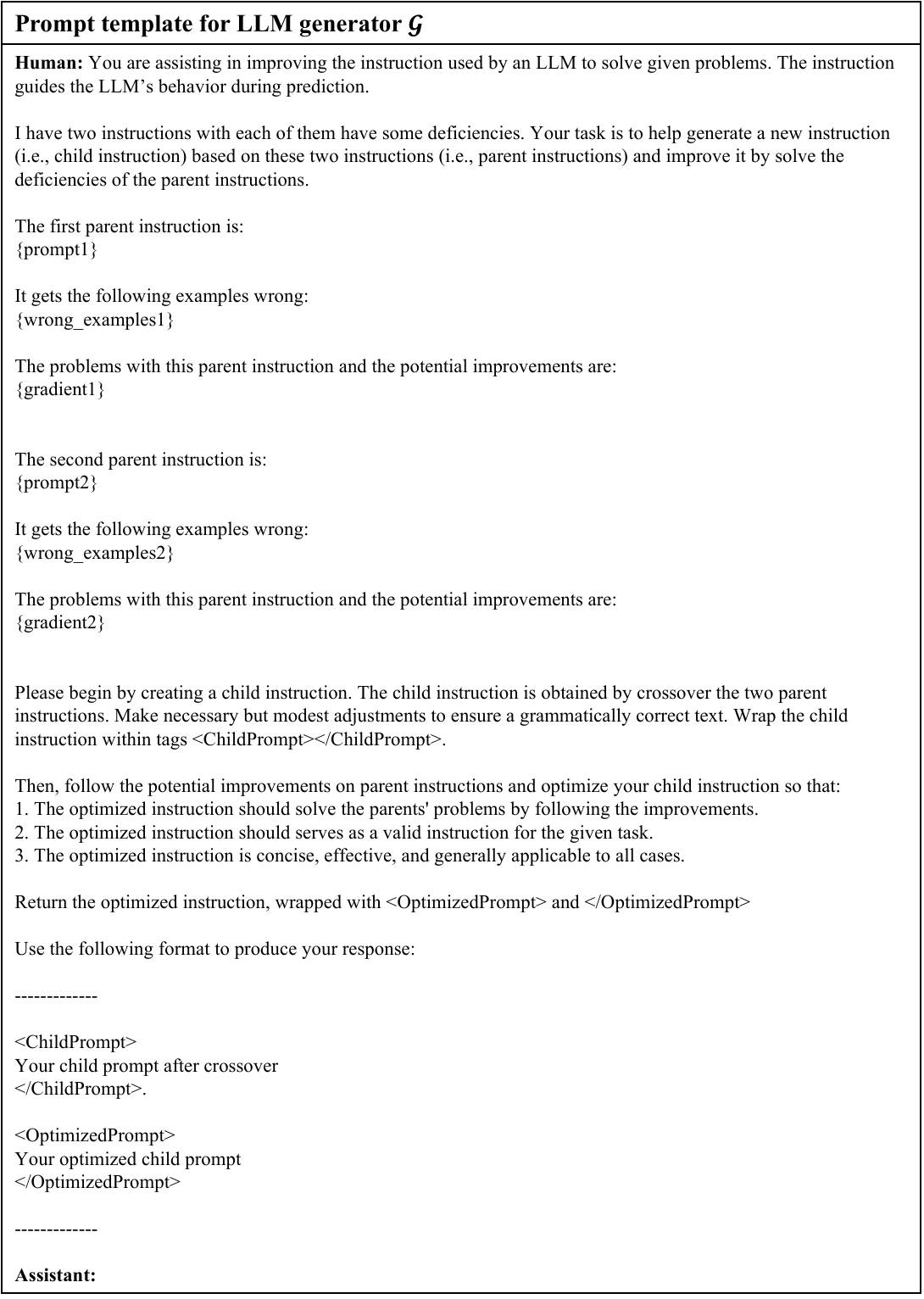}
\end{subfigure}
\caption{Prompt template for generator $\mathcal{G}$}
\label{fig:generator-template}
\end{figure}

\vspace{0.1in}
\noindent \textbf{Meta-prompt template.} See the meta-prompt templates for different agents in \texttt{GreenTEA} in Figure~\ref{fig:predictor-template}, \ref{fig:analyzer-template}, and \ref{fig:generator-template}.

\end{document}